\documentclass[10pt,twocolumn,letterpaper]{article}

\usepackage[pagenumbers]{cvpr} 

\usepackage[dvipsnames]{xcolor}
\usepackage[flushleft]{threeparttable}
\usepackage{colortbl}
\usepackage{multirow}
\usepackage{wrapfig,lipsum,booktabs}
 \usepackage{transparent}

\definecolor{cvprblue}{rgb}{0.21,0.49,0.74}
\usepackage[pagebackref,breaklinks,colorlinks,citecolor=cvprblue]{hyperref}

\def\paperID{656} 
\def\confName{CVPR}
\def\confYear{2024}

\title{InstructSeq: Unifying Vision Tasks with Instruction-conditioned \\ Multi-modal Sequence Generation}

\author{Rongyao Fang$^1$, Shilin Yan$^2$, Zhaoyang Huang$^1$, \\
Jingqiu Zhou$^1$, Hao Tian$^3$, Jifeng Dai$^{4,5}$, Hongsheng Li$^1$\\
$^1$The Chinese University of Hong Kong, $^2$Fudan University,\\
$^3$Sensetime Research Institute, $^4$Tsinghua University, $^5$Shanghai Artificial Intelligence Laboratory\\
{\tt\small rongyaofang@link.cuhk.edu.hk, slyan21@m.fudan.edu.cn,}\\
{\tt\small \{drinkingcoder,1155167063\}@link.cuhk.edu.hk, tianhaohust@gmail.com,} \\
{\tt\small daijifeng@tsinghua.edu.cn, hsli@ee.cuhk.edu.hk}
}

\begin{document}
\maketitle
\begin{abstract}

Empowering models to dynamically accomplish tasks specified through natural language instructions represents a promising path toward more capable and general artificial intelligence. In this work, we introduce InstructSeq, an instruction-conditioned multi-modal modeling framework that unifies diverse vision tasks through flexible natural language control and handling of both visual and textual data. InstructSeq employs a multimodal transformer architecture encompassing visual, language, and sequential modeling. We utilize a visual encoder to extract image features and a text encoder to encode instructions. An autoregressive transformer fuses the representations and generates sequential task outputs. By training with LLM-generated natural language instructions, InstructSeq acquires a strong comprehension of free-form instructions for specifying visual tasks. This provides an intuitive interface for directing capabilities using flexible natural instructions. Without any task-specific tuning, InstructSeq achieves compelling performance on semantic segmentation, referring expression segmentation/comprehension, and image captioning. The flexible control and multi-task unification empower the model with more human-like versatility and generalizability for computer vision. The code will be released soon at \href{https://github.com/rongyaofang/InstructSeq}{https://github.com/rongyaofang/InstructSeq}.
\end{abstract}    
\vspace{-10pt}
\section{Introduction}
\label{sec:intro}

In recent years, the development of multi-modal large language models (MLLMs) using instruction tuning has enabled models to execute tasks based on natural language prompts. Approaches like VisionLLM \cite{wang2023visionllm}, Shikra \cite{chen2023shikra}, and Qwen-VL \cite{bai2023qwen} have demonstrated promising capabilities to accomplish vision-language tasks such as VQA, captioning, and object detection when provided with textual instructions. However, most existing MLLMs tackle various tasks as responses in text format. Therefore, critical vision tasks that require dense outputs like segmentation maps have remained largely unexplored by such MLLM models. This hinders the application of intuitive natural language control to diverse vision tasks requiring pixel-level outputs, posing a key challenge for advancing generalizable multi-modal intelligence.

While recent MLLMs have adopted instruction tuning for natural language control, most prior visual generalist models lack flexible linguistic interfaces. Methods such as Visual Prompt Tuning \cite{jia2022visual} and Painter \cite{wang2023images} adapted vision models to various tasks using visual prompts. However, these frameworks rely on pre-defined prompts to switch between a closed set of capabilities. The lack of free-form natural language instructions makes it challenging to apply these models to novel scenarios specified in unstructured linguistic descriptions. Furthermore, the vision generalist models' exclusive focus on visual inputs and outputs restricts the range of vision-language applications. These models cannot handle textual outputs, constraining their versatility. Establishing unified output formats is crucial for flexibly producing required data types without task-specific modules. This greater generality and flexibility motivate architectures that consolidate diverse vision tasks under both visual and textual outputs within a unified framework. 

On the other hand, approaches like Unified-IO \cite{lu2022unified_io} demonstrate initial capabilities for unifying textual and dense output tasks under natural language control. However, they rely on fixed instruction templates rather than flexible natural language instructions for various tasks. This reliance on restricted instructions causes performance degradation when presented with novel instructions differing from the templates used during training.

We introduce InstructSeq, an instruction-conditioned model that is able to handle both dense and textual output tasks in a unified model, including the tasks of semantic segmentation, reference expression segmentation, reference expression comprehension, and image captioning. InstructSeq leverages both visual and textual modalities for inputs and outputs. It encodes images using a visual encoder and instructions with a pretrained text encoder. The encoded representations are combined and fed to an autoregressive transformer that generates sequences of visual or text tokens based on the input instruction.

Furthermore, considering the limitations of fixed instruction templates, we use an external LLM to create natural language instructions. By handling language more aligned with custom natural descriptions, users can intuitively instruct the model's behavior across tasks rather than relying on restricted instruction templates. We show InstructSeq can learn such alignment by training on rich instruction sets.

Through this flexible multi-modal architecture, InstructSeq achieves substantial improvements over prior work:

\begin{itemize}
    \item It achieves competitive performance across a diverse set of visual and vision-language tasks including semantic segmentation, referring expression segmentation/comprehension, and captioning without task-specific tuning on individual datasets. 

    \item The multi-modal design allows handling both visual and textual inputs and outputs with a unified vison-language autoregressive model, enabling generalizable multi-task capabilities specified by intuitive language instructions.

    \item The instruction set allows the expression of human intentions in various tasks, creating a more generalizable vision system.

\end{itemize}

InstructSeq exhibits an impressive ability to follow free-form natural language instructions for accomplishing a variety of vision tasks. This agility and generalizability, directed by intuitive human-like commands, represents an important step toward artificial general intelligence (AGI) systems for computer vision.

\section{Related Work}
\label{sec:related}

\subsection{Vision Generalist Model}
There has been growing interest in unified models capable of handling diverse vision tasks within a single architecture. Current approaches fall into two categories – language-like sequence generation and image-resembling generation.

Sequence generation methods like OFA \cite{wang2022ofa} and Flamingo \cite{alayrac2022flamingo} frame tasks as text prediction problems. Pix2Seq v2 \cite{chen2022unified_seq} represents spatial outputs like detection using discrete tokens, while Unified-IO \cite{lu2022unified_io} handles dense maps through VQ-VAE \cite{esser2021taming} tokenized outputs. However, these works rely on fixed prompt tuning or fixed language templates rather than flexible natural language instructions.

Image generation approaches like Painter \cite{wang2023images} demonstrate in-context adaptation through masked completion. However, fixed visual contexts limit flexibility and textual handling. PromptDiffusion \cite{wang2023context} enables conditional diffusion for multi-task images but lacks textual outputs. Recently, InstructionDiffusion \cite{geng2023instructdiffusion} and InstructCV \cite{gan2023instructcv} explore expanding language instructions but remain constrained to image outputs by their diffusion architecture.

Our InstructSeq model introduces a sequence generation approach empowered by flexible natural language instructions. Rather than fixed templates, we align free-form linguistic descriptions with model behaviors, achieving superior flexibility. By handling both visual and text, InstructSeq unifies vision and vision-language tasks within a single framework driven by intuitive human-like instructions.

\subsection{Leveraging Natural Language Instructions}

Natural language instructions have emerged as a powerful paradigm for controlling model behavior in both NLP and computer vision. Approaches like InstructGPT \cite{ouyang2022training} and OPT \cite{zhang2022opt} demonstrate enhanced few-shot performance when tuning LLMs on instructional data.

In computer vision, methods like Flamingo \cite{alayrac2022flamingo} and BLIP \cite{li2022blip,li2023blip} connect visual encoders to LLMs through instructional prompting and achieve strong performance on vision-language tasks. However, these models focus on image-to-text problems rather than visual perception tasks.

Some recent works explore visual prompting, like Bar et al.'s token-based image inpainting \cite{bar2022visual} and Painter's masked modeling \cite{wang2023images}. But reliance on fixed visual contexts limits flexibility compared to free-form linguistic instructions. InstructDiffusion \cite{geng2023instructdiffusion} explores expanding natural language instructions for vision tasks, but remains confined to image outputs due to its diffusion-based approach.

Our InstructSeq framework draws inspiration from instruction tuning but differs in its focus on vision tasks and pixel prediction outputs. We align free-form natural language with diverse vision behaviors using a sequence generation approach. Without dependence on fixed visual prompts, InstructSeq leverages rich instruction text to intuitively unify a wide range of vision and vision-language capabilities within a single model.
\section{InstructSeq}
\label{sec:instructseq}

\begin{figure*}
    \centering
    \includegraphics[width=1.0\linewidth]{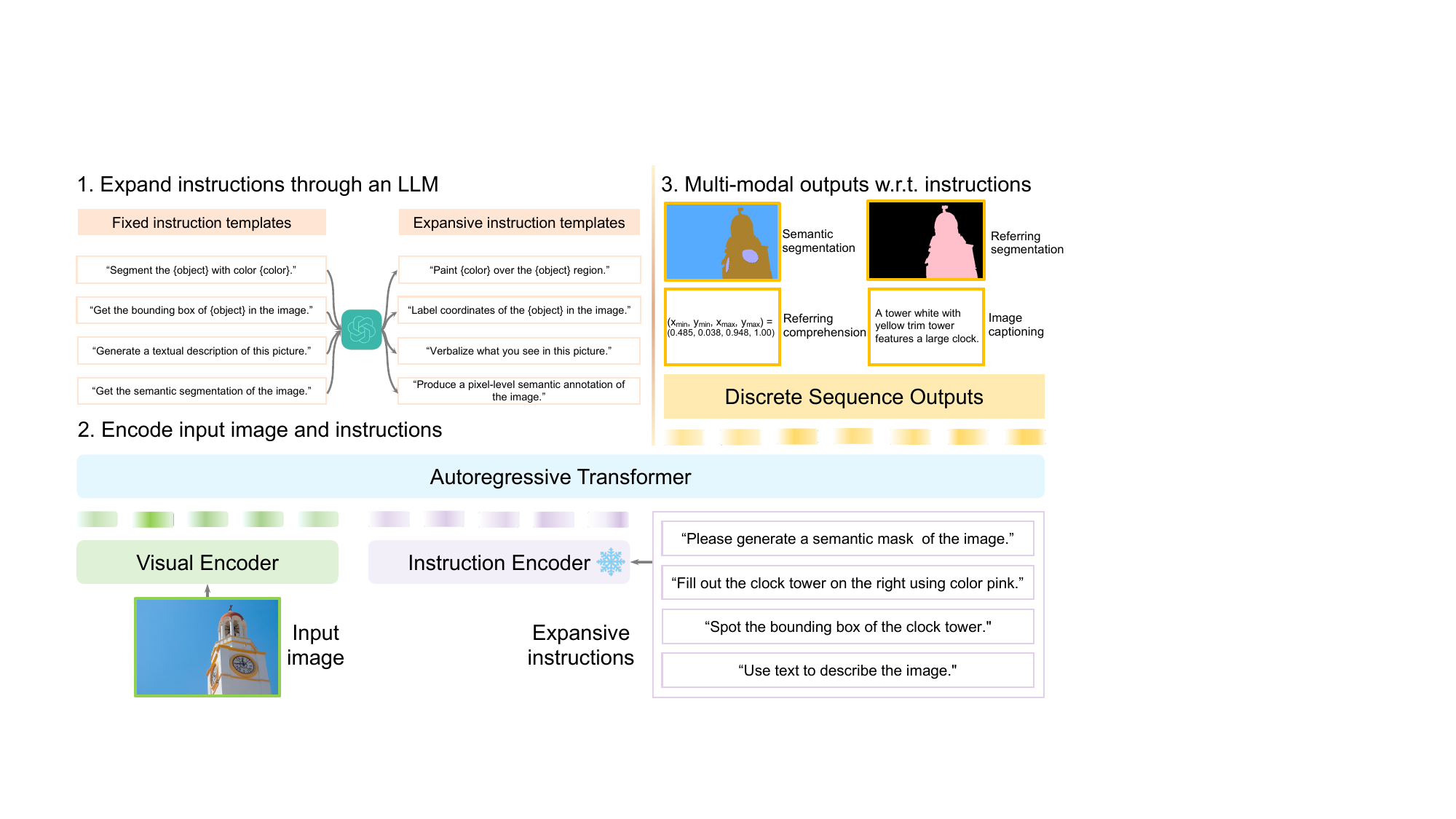}
    \caption{InstructSeq comprises a visual encoder, a frozen instruction encoder, and an autoregressive transformer. The visual encoder extracts image features while the instruction encoder encodes free-form textual instructions. These representations are input to the transformer which generates discrete token sequences. This architecture allows producing various output types to accomplish diverse vision tasks based on natural language directives.}
    \label{fig:main}
\end{figure*}

Our proposed framework, InstructSeq, enables execution of diverse vision tasks through free-form natural language instructions. At a high level, InstructSeq comprises the following stages:

\begin{itemize}
    \item Creating an instruction-task dataset to specify various vision tasks with natural language instruction through an external LLM. The instruction-task pairs are used to train the model to make it follow free-form language instructions to perform various vision tasks.

    \item The input images are encoded through an image encoder and the instructions are encoded by a pretrained frozen language encoder. An autoregressive transformer that generates sequences of output tokens conditioned on the multi-modal inputs. The model can produce both visual and textual sequential outputs to suit different applications.

\end{itemize}

During training, we construct a large multi-task dataset containing images paired with the LLM-generated instructions describing desired tasks including semantic segmentation, reference expression segmentation/comprehension, and captioning. The visual and textual inputs are encoded and passed to the autoregressive transformer, which is trained end-to-end to generate appropriate output token sequences based on the input instructions.

During inference, InstructSeq takes an image and free-form natural language describing the task objective and then predicts an output sequence of tokens corresponding to the instructed task, which is decoded into the final output. 

This approach of combining flexible language instructions, multi-modal input representations, and sequence generation provides a flexible framework for unifying diverse vision capabilities. InstructSeq handles both visual and textual inputs and outputs, directed entirely through intuitive natural language instructions.

\subsection{Unified Instructions for Multiple Vision Tasks}

InstructSeq takes natural language instructions to specify model behavior across diverse vision tasks. To enable flexible task selection, we use an external LLM to generate rich textual instructions covering a wide range of natural phrasings for specifying tasks.

We leverage the Anthropic Claude-2 chatbot \cite{claude}, a large conversational model. We provide Claude-2 with template instructions for vision tasks such as \textit{``Please segment the \{object\} with color \{color\}."} or \textit{``Provide me with the caption of this image."} Claude-2 then outputs paraphrased natural language instructions for accomplishing these tasks that align with human notions of commanding visual perception. 

This instruction set allows our model to intuitively specify tasks through natural phrasings rather than rigid templates. The generated instructions are encoded using a frozen pretrained text encoder.

\subsection{Multi-modal Sequence Generation Framework}

\subsubsection{Architecture}

The InstructSeq model comprises three key components to enable instruction-conditioned sequence generation as shown in \cref{fig:main}:

\noindent \textbf{Visual Encoder.} We utilize a Vision Transformer (ViT) \cite{dosovitskiy2020image} to encode visual features from input images. Input images are inputted at resolution $512\times512$. The ViT generates 1024 image token embeddings of $16\times16$ patches summarizing the visual content. After comparing encoders including MAE \cite{he2022masked}, SAM \cite{kirillov2023segment} and CLIP \cite{radford2021learning}, we empirically select MAE due to its strong performance on our downstream tasks.

\noindent \textbf{Instruction Encoder.} For encoding instructions, we employ RoBERTa \cite{liu2019roberta} as the instruction encoder. RoBERTa encodes the natural language instructions into vector representations capturing the linguistic meanings and intents. The text encoder parameters are frozen during training.

\noindent \textbf{Autoregressive Transformer.} The 1024 image tokens and text tokens are projected to a common embedding space and concatenated at the sequence level to form the transformer input. It is used as input to an autoregressive transformer. The transformer is trained end-to-end to generate output sequences of tokens conditioned on the joint input embedding. The model predicts tokens from a unified vocabulary containing separate sets of visual, positional, and text tokens. Given the instruction, the model adaptively produces tokens from the corresponding vocabulary set.

\subsubsection{Task Formulations and Data Constructions}

InstructSeq produces three types of outputs using different formulations - dense visual tokens, sparse positional tokens, and text tokens.

\begin{itemize}
    \item \textbf{Dense visual tokens.} It represents pixel predictions for full image vision tasks like semantic segmentation and referring segmentation. We leverage VQ-VAE \cite{esser2021taming} to vector quantize dense target images into discrete visual tokens. The indices of VQ tokens are mapped to a specific vocabulary set in the autoregressive transformer.

    \item \textbf{Sparse positional tokens.} It represents discretized image coordinates for the referring comprehension (detection) task. We utilize an additional set of 1000 positional tokens to represent coordinate locations in the autoregressive transformer. Each bounding box in the detection task is identified by a sequence of 4 positional tokens denoting its $[x1, y1, x2, y2]$ coordinates.

    \item \textbf{Text tokens.} Text outputs are tokenized using a 50k BPE vocabulary. This allows for generating free-form language for tasks like captioning.

\end{itemize}

We construct the dataset from publicly available datasets for the three types of output tokens.

\noindent \textbf{Semantic segmentation} is a dense prediction task that assigns a semantic label to each pixel in an input image. We utilize the datasets COCO-Stuff \cite{caesar2018coco} and ADE20K \cite{zhou2017scene} which provide pixel-level semantic labels for images across categories. We map each of the classes to a distinct color, evenly spaced across RGB space. This creates a visual segmentation image where each colored pixel represents a semantic class. We vector quantize the visual segmentation image into discrete visual tokens using the pretrained VQ-VAE encoder. At inference time, the model's predicted visual tokens are decoded by the VQ-VAE decoder to reconstruct the visual segmentation image. The image is then discretized to the nearest class-color mapping to obtain the final per-category segmented output. For the semantic segmentation task, we utilize the instructions like \textit{``Get the semantic segmentation in COCO Stuff format."} or \textit{``Generate a semantic mask in ADE20K style."}.

\noindent \textbf{Reference expression segmentation} localizes an object segmentation mask within an image based on a natural language referring expression. We utilize datasets including RefCOCO, RefCOCO+, RefCOCOg \cite{yu2016refcoco}, and gRefCOCO \cite{liu2023gres} which contain images paired with referring expressions about particular objects.

We create instructions like \textit{``Please segment the \{object\} with color \{color\}."} For each sample, we randomly select a color and replace \textit{``\{object\}"} with the referring expression. The target segmentation map is generated by masking the referred object region with the chosen color. The segmentation map image is tokenized by the VQ-VAE and used as the training objective.

At inference time, InstructSeq takes the image and the instruction containing a natural referring expression as input. Then it produces a sequence of visual tokens to reconstruct back to the segmentation map corresponding to the referring object. 

\noindent \textbf{Reference expression comprehension (detection)} localizes the bounding box of an object within an image based on a natural language expression. We utilize datasets of RefCOCO, RefCOCO+, and RefCOCOg \cite{yu2016refcoco}. We represent each bounding box using a sequence of 4 positional tokens indicating $[x1, y1, x2, y2]$ coordinates of the top-left and bottom-right corners. The positional tokens are identified by $1000$ tokens in the autoregressive transformer. At inference time, InstructSeq takes as input an instruction including referring expression, and produces a sequence of positional tokens. The instructions for this task are like \textit{``Get the bounding box of \{object\} in the image."}, where the \textit{``\{object\}"} is replaced by the referring expression.

\noindent \textbf{Image captioning} involves generating a natural language description for a given image. We use the COCO Captions \cite{chen2015cococaption} dataset which provides human-written captions describing the content of images. We encode the caption into sequences of subword tokens using byte-pair encoding with a vocabulary size of $50,265$. This enables formulating caption generation as a text sequence prediction task. InstructSeq is trained to produce sequences of text tokens when provided an instruction like \textit{``Generate a textual description of this picture."}.

To construct the full training data, we sample image-instruction-target triplets from the various task datasets described above using a fixed sampling ratio. For each training instance, an instruction is randomly selected from the instruction set for that task and paired with the corresponding input-output data. 

\vspace{-10pt}

\subsubsection{Training Objective}
We optimize the model using a token-wise cross-entropy loss between the predicted token probabilities and ground truth target tokens. By training on varied multi-modal data spanning diverse tasks, this consistent loss trains the unified architecture to follow instructions and generate suitable outputs across multiple vision and language applications.

\section{Experiments}
\label{sec:exp}

\begin{table*}[tb!]
  \centering
  \setlength{\tabcolsep}{0.8mm}{
  \begin{tabular}{l|cccccccccccc}
    \toprule
    \multirow{3}{*}{} &
    \multicolumn{2}{c}{\textbf{Semantic Segmentation}} & \multicolumn{2}{c}{\textbf{Referring Segmentation}} &
    \multicolumn{1}{c}{\textbf{Referring Comprehension}} &
    \textbf{Image Captioning} \\
    & \multicolumn{2}{c}{mIoU$\uparrow$} & \multicolumn{2}{c}{oIoU$\uparrow$} & AP50$\uparrow$ & BLEU$\uparrow$ \\
    & COCO-Stuff & ADE20K & RefCOCO & gRefCOCO & RefCOCO  & COCO-Captions &  \\
    \toprule
    \multicolumn{8}{c}{\textbf{Specialised models}} \\
    \midrule
    SSA \cite{chen2023semantic} & * & 47.15 & \cellcolor[gray]{0.9} & \cellcolor[gray]{0.9} & \cellcolor[gray]{0.9} & \cellcolor[gray]{0.9} \\
    SegFormer \cite{xie2021segformer} & 46.70 & 51.80 & \cellcolor[gray]{0.9} & \cellcolor[gray]{0.9} & \cellcolor[gray]{0.9} & \cellcolor[gray]{0.9}  \\
    Mask2Former \cite{cheng2022mask2former} & * & 56.10 & \cellcolor[gray]{0.9} & \cellcolor[gray]{0.9} & \cellcolor[gray]{0.9} & \cellcolor[gray]{0.9} \\
    CMSA \cite{ye2019cross} & \cellcolor[gray]{0.9} & \cellcolor[gray]{0.9} & 58.32 & * & \cellcolor[gray]{0.9} & \cellcolor[gray]{0.9} \\
    LAVT \cite{yang2022lavt} & \cellcolor[gray]{0.9} & \cellcolor[gray]{0.9} & 72.73 & 57.64 & \cellcolor[gray]{0.9} & \cellcolor[gray]{0.9} \\
    ReLA \cite{liu2023gres} & \cellcolor[gray]{0.9} & \cellcolor[gray]{0.9}  & 73.21 & 62.42 & \cellcolor[gray]{0.9} & \cellcolor[gray]{0.9} \\
    UNITER \cite{chen2020uniter} & \cellcolor[gray]{0.9} & \cellcolor[gray]{0.9}   & \cellcolor[gray]{0.9} & \cellcolor[gray]{0.9} & 81.41 & \cellcolor[gray]{0.9} \\
    MDETR \cite{kamath2021mdetr} & \cellcolor[gray]{0.9} & \cellcolor[gray]{0.9}  & \cellcolor[gray]{0.9} & \cellcolor[gray]{0.9} & 86.75 & \cellcolor[gray]{0.9} \\
    \toprule
    \multicolumn{8}{c}{\textbf{Generalist models}} \\
    \midrule
    Unified-IO \cite{lu2022unified_io} & * & 25.71 & 46.42 & 17.31 & * & *\\
    Pix2SeqV2 \cite{chen2022unified_seq} & \cellcolor[gray]{0.9} & \cellcolor[gray]{0.9} & \cellcolor[gray]{0.9} & \cellcolor[gray]{0.9}  & \cellcolor[gray]{0.9} & 34.90 \\
    InstructCV \cite{gan2023instructcv} & * & 47.24 & \cellcolor[gray]{0.9} & \cellcolor[gray]{0.9} & \cellcolor[gray]{0.9} & \cellcolor[gray]{0.9} \\
    InstructSeq & 47.22 & 47.01 & 59.60 & 61.69 & 75.22 & 35.00 \\
    \bottomrule
  \end{tabular}}
  \caption{Performance comparison of InstructSeq with specialised models and other generalist models. \colorbox{gray!20}{The gray cells} denote the model cannot handle the task. * denotes the model is capable of the task but is not evaluated on the benchmark.}\label{table:result}
\end{table*}

\subsection{Settings}

\textbf{Datasets.} For the detailed introduction of the datasets ADE20K \cite{zhou2017scene}, COCO-Stuff \cite{caesar2018coco}, RefCOCO, RefCOCO+, RefCOCOg \cite{yu2016refcoco}, gRefCOCO \cite{liu2023gres}, and COCO Captions \cite{chen2015cococaption}.

\noindent \textbf{Evaluation metrics.} For semantic segmentation, we adopt the mean Intersection-over-Union (mIoU) metric. RES is evaluated by the overall Intersection-over-Union (oIoU) metric between predicted and true segmentation masks. For REC, we report AP50 accuracy in predicting referred object boxes. For image captioning, we report the BLEU score.

\noindent \textbf{Data sampling ratio.} During training, we sample data from the various tasks using a fixed ratio. The sampling weight for each task is 0.25 for COCO-Stuff (semantic segmentation), 0.2 for ADE20K (semantic segmentation), 0.05 each for RefCOCO/RefCOCO+/RefCOCOg/gRefCOCO (RES), 0.05 each for RefCOCO/RefCOCO+/RefCOCOg (REC), and 0.1 for COCO Captions (image captioning). 

\noindent \textbf{Model scale.} InstructSeq uses MAE$_{\text{\ LARGE}}$ \cite{he2022masked} as the visual encoder and RoBERTa$_{\text{\ LARGE}}$ \cite{liu2019roberta} as the text encoder. The autoregressive transformer has $12$ layers, with embedding size $1024$ and head number $16$. The visual encoder and autoregressive transformer are trainable, while the text encoder is frozen.

\noindent \textbf{Implementation details.} InstructSeq takes $512\times512$ pixel images as input. The VQ-VAE \cite{esser2021taming} decoder reconstructs $256\times256$ output images, with a downsampling factor of $8\times8$. We initialize the VQ-VAE weights from a pretrained checkpoint \cite{rombach2022high}, then finetune using our output images.

For evaluation across tasks except image captioning, we generate tokens by sampling from the predicted distribution with a temperature between $0.8$ and $1.0$. The optimal temperature varies slightly across tasks. We also sample $10$ output sequences per input and aggregate predictions, leveraging InstructSeq's sampling-based capabilities as discussed in \cref{sec:sampling}. For image captioning, we apply beam search with a beam size of $6$ to generate caption texts.

\begin{table*}[tb!]
  \centering
  \setlength{\tabcolsep}{5.0mm}{
  \begin{tabular}{lccc}
    \toprule
    \multirow{2}{*}{Instruction} & InstructSeq-Template & InstructSeq-Full \\
    \cline{2-3}
    &  \multicolumn{2}{c}{oIoU}   \\
    \midrule
    \rowcolor{gray!15} Instruction template: \textit{``Segment the \{object\} with color \{color\}."} &  59.32 & 59.60  \\
    \textit{``Mark the \{object\} region as \{color\}."} & 42.10 & 57.95 \\
    \textit{``Please fill \{color\} into the shape of \{object\}."} & 45.65 & 59.54 \\
    \textit{``Specify the \{object\} by painting it \{color\}."} & 49.85 & 58.40 \\
    \textit{``View the \{object\} as \{color\}."} & 30.87 & 59.43 \\

    \bottomrule
  \end{tabular}}
  \caption{Evaluation InstructSeq model with new written instructions on the RefCOCO referring segmentation task. The InstructSeq-Template model is trained with a fixed instruction template for each task, while our base setting InstructSeq-Full is trained with the LLM expanded instruction set. The new written instructions are not included in both model's training sets.}\label{table:flexible}

  \vspace{-10pt}
\end{table*}

\subsection{Results}

\textbf{Performance comparisons.} We evaluate InstructSeq on four representative vision and vision-language tasks: semantic segmentation, referring expression segmentation, referring expression comprehension, and image captioning. The performance compared to prior specialised and generalist models is shown in \cref{table:result}.

For semantic segmentation on COCO-Stuff \cite{caesar2018coco} and ADE20K \cite{zhou2017scene}, InstructSeq achieves competitive mIoU scores of 47.22 and 47.01 respectively. On COCO-Stuff, it outperforms the specialised SegFormer \cite{xie2021segformer} by 0.52 mIoU. On ADE20K, InstructSeq surpasses the generalist Unified-IO \cite{lu2022unified_io} by a large margin of 20.53 mIoU and is also comparable to the recent work InstructCV \cite{gan2023instructcv}. These results validate InstructSeq's strong segmentation capabilities.

On referring segmentation, InstructSeq attains significantly higher oIoU of 59.60 on RefCOCO and 61.69 on gRefCOCO compared to Unified-IO \cite{lu2022unified_io}. It also matches the performance of prior specialised models. This demonstrates InstructSeq's contextual understanding of localizing objects based on natural language expressions.

For referring comprehension on RefCOCO, InstructSeq achieves 75.22 AP50 without relying on bounding box supervision but sparse token supervision during training.

On COCO Captions for image captioning, InstructSeq obtains a BLEU score of 35.00, slightly outperforming the generalist Pix2SeqV2 \cite{chen2022unified_seq} model.

The strong well-rounded performance confirms InstructSeq's generalizability to diverse vision and vision-language tasks under flexible natural language control. 

\noindent \textbf{Generalization to new instructions.} A key capability of InstructSeq is its ability to generalize to new instructions beyond those seen during training. We validate this by comparing it to a baseline InstructSeq-Template model, which is trained only on fixed instruction templates for each task.

As shown in \cref{table:flexible}, when evaluated on novel instructions for referring segmentation on RefCOCO, our full InstructSeq model achieves consistent performance within 2\% oIoU of its original scores. However, InstructSeq-Template suffers dramatic degradation of over 10\% oIoU on novel phrasings, and even 30\% drop on the instruction \textit{``View the \{object\} as \{color\}.”}

\begin{figure*}
    \centering
    \includegraphics[width=1.0\linewidth]{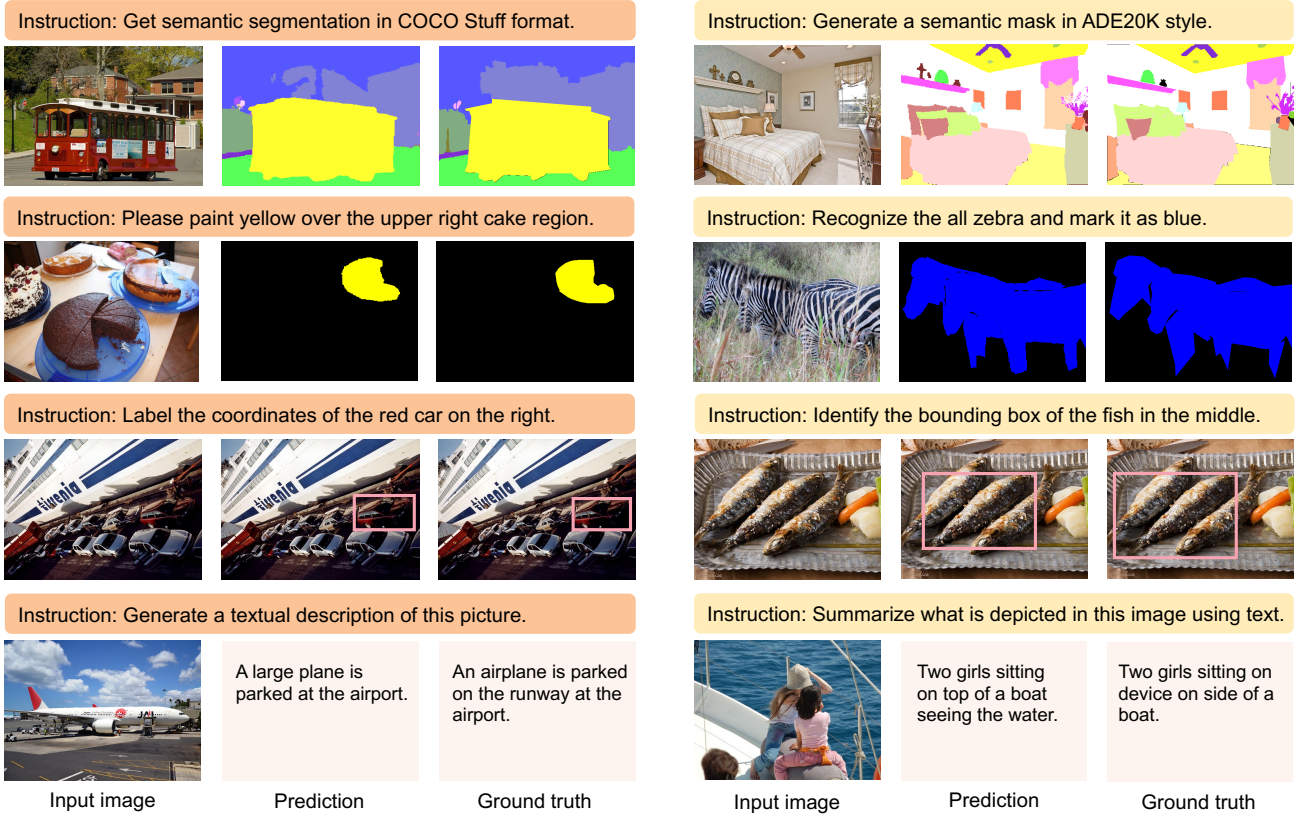}
    \caption{Qualitative results of InstructSeq across all vision tasks.}
    \label{fig:qualitative}
    \vspace{-10pt}
\end{figure*}

This demonstrates InstructSeq's strong generalization beyond its training instruction set. By training on the instruction set from an LLM, InstructSeq acquires robust language understanding and alignment with natural phrasings. The flexible formulation frees it from overfitting to templates. In contrast, constraining language to predetermined formats limits generalization capabilities, as evidenced by InstructSeq-Template's inability to comprehend new instructions.

\begin{wraptable}{r}{0.18\textwidth}
\vspace{-10pt}
\hspace{-10pt}
    \begin{minipage}[t]{0.18\textwidth}
    \centering
    
    \label{table:unseen}
	\setlength{\tabcolsep}{0.7mm}{\begin{tabular}{lc}
	\toprule
        Model & mIoU$\uparrow$ \\
        \midrule
        Inpainting \cite{bar2022visual} & 58.5   \\
        Painter \cite{wang2023images} & 62.3  \\        
        InstructSeq & \textbf{68.7} \\
	\bottomrule
	\end{tabular}}
 \caption{Evaluation of InstructSeq's segmentation ability on FSS-1000 \cite{li2020fss} benchmark.}
    \end{minipage}
\end{wraptable}

\noindent \textbf{Generalization to unseen classes.} Since our model is trained with natural language instructions with the implementation of a pretrained instruction encoder. It is expected the model has the ability to unseen classes. To assess InstructSeq's ability to generalize to unseen classes beyond those observed during training, we evaluate it on the 1000-class FSS-1000 segmentation benchmark \cite{li2020fss}. This dataset comprises $1000$ distinct categories which are mostly not present in our training set. We apply our InstructSeq model to segment objects based on instructions specifying unseen classes. InstructSeq achieves a strong mIoU of 68.7\% on this out-of-distribution data. For comparison, the Painter \cite{wang2023images} attains 62.3\% mIoU, while Inpainting \cite{bar2022visual} yields 58.5\% mIoU. The considerable gains over both approaches exemplify InstructSeq's ability to flexibly apply learned segmentation skills to novel unseen categories specified through intuitive natural language commands. The results validate its potential for few-shot learning and adapting to open-vocabulary categories beyond those in fixed pre-training datasets.

\noindent \textbf{Qualitative results.} To intuitively exhibit InstructSeq's capabilities on diverse tasks, we present the qualitative result figure covering examples from all applications.

As shown in \cref{fig:qualitative}, the top row displays InstructSeq's semantic segmentation predictions on sample COCO-Stuff and ADE20K images. The model accurately segments small objects while maintaining coherent regions. The second row shows reference expression segmentation results. InstructSeq localizes referred objects with precise masks based on the natural language expressions. The third row demonstrates referring expression comprehension, where InstructSeq grounds objects in bounding boxes using only sparse token supervision. Finally, the bottom row provides image captions, covering diverse descriptive content.

\subsection{Leveraging Sampling-Based Prediction} \label{sec:sampling}

Unlike prior deterministic vision models that produce fixed outputs, InstructSeq performs sampling-based prediction by generating tokens according to their probability distribution. This difference enables new capabilities compared to deterministic models. By sampling multiple token sequences, InstructSeq can produce varied outputs for the same input. It also provides confidence scores for each generated token based on their probabilities. We analyze the benefits of such sampling-based prediction on vision tasks.

\noindent \textbf{Multiple sampling for better performance.} Sampling multiple sequences during inference allows aggregating predictions to improve accuracy. 

For semantic segmentation, we sample multiple token sequences to produce $N$ distinct segmentation maps per image. We then aggregate the predictions by voting the class with maximum frequency at each pixel location across the $N$ samples.

\begin{table}[tb!]
  \centering
  \setlength{\tabcolsep}{0.8mm}{
  \begin{tabular}{c|cccccccccc}
    \toprule
    \multirow{2}{*}{Num.} & COCO-Stuff & ADE20K & RefCOCO & RefCOCO   \\
    & mIoU & mIoU & oIoU (RES) & AP50 (REC)   \\
    \midrule
    1 &  40.72 & 42.54 & 53.07 & 70.71\\
    4  & 44.40 & 46.24 & 57.59 & 72.30\\
    6  & 46.09 & 46.31 & 58.01 & 73.93\\
    8  & 46.88 & 46.52 & 58.94 & 74.28\\
    10  & 47.22 & 47.01 & 59.60 & 75.22\\

    \bottomrule
  \end{tabular}}
  \caption{Performance of InstructSeq sampling multiple numbers during evaluation on each task.}\label{table:sampling}

  \vspace{-10pt}
\end{table}

\begin{table*}[tb!]
  \centering
  \setlength{\tabcolsep}{1.2mm}{
  \begin{tabular}{cc|ccccccccccccc}
    \toprule
    \multirow{2}{*}{Visual Encoder} & \multirow{2}{*}{Text Encoder} & COCO-Stuff & ADE20K & RefCOCO (RES) & RefCOCO (REC)  & COCO-Captions &  \\
    & & mIoU & mIoU & oIoU & AP50  & BLEU &  \\
    \midrule
    MAE$_{\text{\ BASE}}$ \cite{he2022masked} & RoBERTa$_{\text{\ BASE}}$ \cite{liu2019roberta} & \textbf{42.77} & \textbf{42.13} & 50.41 & \textbf{59.01} & 25.10  \\
    SAM$_{\text{\ BASE}}$ \cite{kirillov2023segment} & RoBERTa$_{\text{\ BASE}}$ \cite{liu2019roberta} & 40.86 & 41.63 & 50.09 & 58.61 & \textbf{26.90} \\
    CLIP$_{\text{\ BASE}}$ \cite{radford2021learning} & CLIP$_{\text{\ BASE}}$ \cite{radford2021learning} & 42.55 & 42.10 & \textbf{51.12} & 56.89 & 25.50 \\

    \bottomrule
  \end{tabular}}
  \caption{Performance comparison of InstructSeq with specialised models and other generalist models.}\label{table:encoder}
  \vspace{-7pt}
\end{table*}

Similarly, for referring expression segmentation, we generate $N$ samples of segmentation masks for each expression and select the mask with the highest intersection-over-union (IoU) with the other predicted samples, which is typically the most accurate. For referring expression comprehension, we sample $N$ bounding box predictions per expression and choose the box with maximum IoU against the other predicted samples. As shown in \cref{table:sampling}, the performances increase steadily with more sampling and aggregation, demonstrating the utility of sampling-based prediction.

\begin{figure}
\vspace{-3pt}
    \centering
    \includegraphics[width=1.0\linewidth]{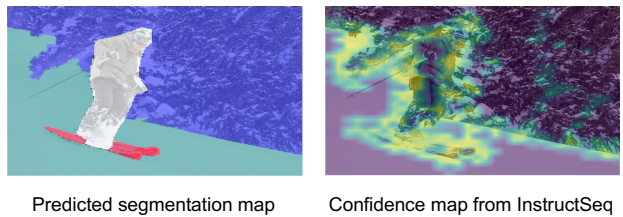}
    \caption{\textbf{Left:} semantic segmentation map generated from the InstructSeq model. \textbf{Right:} confidence map obtained during InstructSeq token sampling. The yellow areas denote low-confidence areas and the purple areas denote high-confidence areas.}
    \vspace{-12pt}
    \label{fig:confidence}
\end{figure}

The sampling-based prediction allows the production of varied outputs to mitigate noise and uncertainty. Aggregating multiple samples improves robustness and enhances performance.

\noindent \textbf{Providing confidence map for dense prediction.} In addition to sampling multiple outputs, the token probabilities provided by InstructSeq during generation can be utilized to estimate prediction confidence. For dense vision tasks like segmentation, we can construct a spatial confidence map by arranging the token probabilities into a spatial grid corresponding to the output dimensions.

Areas of high-probability tokens will result in high confidence scores on the map. Similarly, low-probability tokens translate to regions of uncertainty. The example in \cref{fig:confidence} illustrates InstructSeq's ability to predict segmentation confidence maps by compiling token probabilities.

The model expresses lower confidence around object boundaries and small regions, while large contiguous areas obtain high confidence, accurately reflecting uncertainty. By capturing variance in token probabilities, InstructSeq identifies reliable areas in its stochastic outputs.

The confidence estimation complements the accuracy improvements from multi-sampling aggregation. Together, these properties enhance InstructSeq's reliability compared to fixed deterministic predictions. The model's awareness of its own certainty is valuable for practical applications. The stochastic formulation provides richer information to understand and interpret the model's prediction process.

\begin{table}[tb!]
  \centering
  \setlength{\tabcolsep}{1.2mm}{
  \begin{tabular}{c|cccccccccc}
    \toprule
    \multirow{2}{*}{Output type} & COCO-Stuff & ADE20K & RefCOCO   \\
    & mIoU & mIoU & oIoU   \\
    \midrule
    All tasks  & \textbf{42.77} &  42.13 & 50.41 \\
    Image-output only  & 41.99 & \textbf{42.65} & \textbf{56.39}  \\

    \bottomrule
  \end{tabular}}
  \caption{Ablation on InstructSeq$_{\text{\ BASE}}$ training with all tasks and training with image-output tasks only.}\label{table:mix}
  \vspace{-12pt}
\end{table}

\vspace{-3pt}
\subsection{Ablation Studies} \label{sec:ablation}
\vspace{-2pt}

We ablate InstructSeq's performance on encoder selection and the influence of mixing output types. Due to computational resource constraints, we implement the ablation with a BASE scale model, which uses MAE$_{\text{\ BASE}}$ as the visual encoder and RoBERTa$_{\text{\ BASE}}$ as the text encoder with a 12-layer autoregressive transformer.

\textbf{Ablation on encoder selection.} We analyze the impact of different encoder choices on downstream task performance using base-scale InstructSeq variants.

The default configuration utilizes MAE \cite{he2022masked} for visual encoding and RoBERTa \cite{liu2019roberta} for instruction encoding. We compare with (1) replacing the visual encoder with SAM \cite{kirillov2023segment}, and (2) replacing both encoders with CLIP \cite{radford2021learning}. Quantitative results are shown in \cref{table:encoder}.

Overall, the MAE visual encoder paired with the RoBERTa text encoder achieves the best performance on dense prediction tasks like segmentation and sparse localization tasks like REC. For image captioning, combining SAM and RoBERTa performs the best. Surprisingly, SAM underperforms MAE on most tasks despite strong specialised performance on segmentation. A potential reason is the discrepancy between SAM's DETR-liked architecture and the VQ-VAE representation used in InstructSeq.

\noindent \textbf{Ablation on mixing output types.} InstructSeq trains on datasets covering dense images, sparse positional, and textual outputs. We analyze the impact of mixing textual and dense visual outputs via an ablation.

Specifically, we train a variant using only dense image output datasets, removing textual REC and captioning. As shown in \cref{table:mix}, both models achieve similar segmentation performance on COCO-Stuff and ADE20K. However, the image-output only model attains higher accuracy on RefCOCO referring segmentation.

The performance differences suggest mixing output types does not substantially affect semantic segmentation, but harms referring segmentation. A potential reason is the instruction overlap between referring segmentation and comprehension - both include referring expressions. This may influence the understanding of the referring segmentation instructions when trained jointly with comprehension, degrading segmentation accuracy.

However, the minor COCO-Stuff and ADE20K differences also suggest mixing diverse outputs does not directly harm performance. InstructSeq can still learn effectively across heterogeneous outputs while delivering generalizability benefits over specialised models. Supporting varied data types remains critical for flexible applicability to both visual and textual tasks.

In summary, while differences exist across output types, InstructSeq maintains strong capabilities when trained on mixed outputs. The unified formulation provides generalizability with minor trade-offs compared to isolated training.

\vspace{-5pt}
\section{Conclusion}
\label{sec:conclusion}
\vspace{-3pt}

We presented InstructSeq, a unified multi-task vision model directed by natural language instructions. InstructSeq combines multi-modal encoders with a sequential transformer to generate pixel or text outputs conditioned on input images and free-form textual instructions.

By training on an expanded instruction set from LLM, InstructSeq achieves strong alignment with human notions of task direction. Our model accomplishes a diverse set of segmentation, detection, and captioning tasks under flexible natural language control, rivaling specialised models without task-specific tuning.

InstructSeq also demonstrates compelling generalization capabilities to novel data through its instruction-following formulation. The sampling-based prediction further provides benefits like confidence estimation.

Overall, the natural language-conditioned formulation represents a step towards more human-aligned AI systems. By unifying vision objectives through intuitive natural language commands, we aim to move closer to AGI.

{
    \small
    \bibliographystyle{ieeenat_fullname}
    \bibliography{main}
}


\end{document}